\documentclass[10pt,twocolumn,letterpaper]{article}

\usepackage{cvpr}
\usepackage{times}
\usepackage{epsfig}
\usepackage{graphicx}
\usepackage{amsmath}
\usepackage{amssymb}
\usepackage{subcaption}
\usepackage{algorithm}
\usepackage{algorithmic}
\usepackage{booktabs}
\usepackage{subcaption}

\usepackage[pagebackref=true,breaklinks=true,letterpaper=true,colorlinks,bookmarks=false]{hyperref}

\cvprfinalcopy % *** Uncomment this line for the final submission

\def\cvprPaperID{5433} % *** Enter the CVPR Paper ID here

\ifcvprfinal\fi
\begin{document}

%%%%%%%%% TITLE
\title{AETv2: AutoEncoding Transformations for Self-Supervised Representation Learning by Minimizing Geodesic Distances in Lie Groups}

\author{Feng Lin~$^{1}$, Haohang Xu~$^{2}$, Houqiang Li~$^{1}$, Hongkai Xiong~$^{2}$, {\bf Guo-Jun Qi}~$^{3,}$\thanks{Corresponding author: Guo-Jun Qi, e-mail: guojunq@gmail.com. Guo-Jun Qi conceived the idea, formulated the method, and prepared the paper, while Feng Lin and Haohang Xu performed experiments on the Huawei Cloud.}\vspace{1mm}\\
$^1$University of Science and Technology of China,~$^2$Shanghai Jiao Tong University\\
$^3$ Huawei Research America \vspace{1.5mm}\\
Laboratory for MAchine Perception and LEarning (MAPLE)\\
%\url{http://maple-lab.net/}\vspace{1.5mm}\\
%{\tt\small guojun.qi@huawei.com}\\
{\small\url{http://maple-lab.net/projects/AETv2.htm}}
%Bellevue, WA\\
%{\tt\small guojunq@gmail.com}
% For a paper whose authors are all at the same institution,
% omit the following lines up until the closing ``}''.
% Additional authors and addresses can be added with ``\and'',
% just like the second author.
% To save space, use either the email address or home page, not both
%\and
%Second Author\\
%Institution2\\
%First line of institution2 address\\
%{\tt\small secondauthor@i2.org}
}

\maketitle
%\thispagestyle{empty}

%%%%%%%%% ABSTRACT
\begin{abstract}
Self-supervised learning by predicting transformations has demonstrated outstanding performances in both unsupervised and (semi-)supervised tasks. Among the state-of-the-art methods is the AutoEncoding Transformations (AET) by decoding transformations from the learned representations of original and transformed images. Both deterministic and probabilistic AETs rely on the Euclidean distance to measure the deviation of estimated transformations from their groundtruth counterparts. However, this assumption is questionable as a group of transformations often reside on a curved manifold rather staying in a flat Euclidean space. For this reason, we should use the geodesic to characterize how an image transform along the manifold of a transformation group, and adopt its length to measure the deviation between transformations.  Particularly, we present to autoencode a Lie group of homography transformations ${\bf PG}(2)$  to learn image representations. For this, we make an estimate of the intractable Riemannian logarithm by projecting ${\bf PG}(2)$ to a subgroup of rotation transformations ${\bf SO}(3)$ that allows the closed-form expression of geodesic distances. Experiments demonstrate the proposed AETv2 model outperforms the previous version as well as the other state-of-the-art self-supervised models in multiple tasks.
\end{abstract}

%%%%%%%%% BODY TEXT
\section{Introduction}

\begin{figure}[t]
%\captionsetup[subfigure]{singlelinecheck=false}
    \centering
    \begin{subfigure}[c]{0.5\textwidth}
        \includegraphics[width=\textwidth]{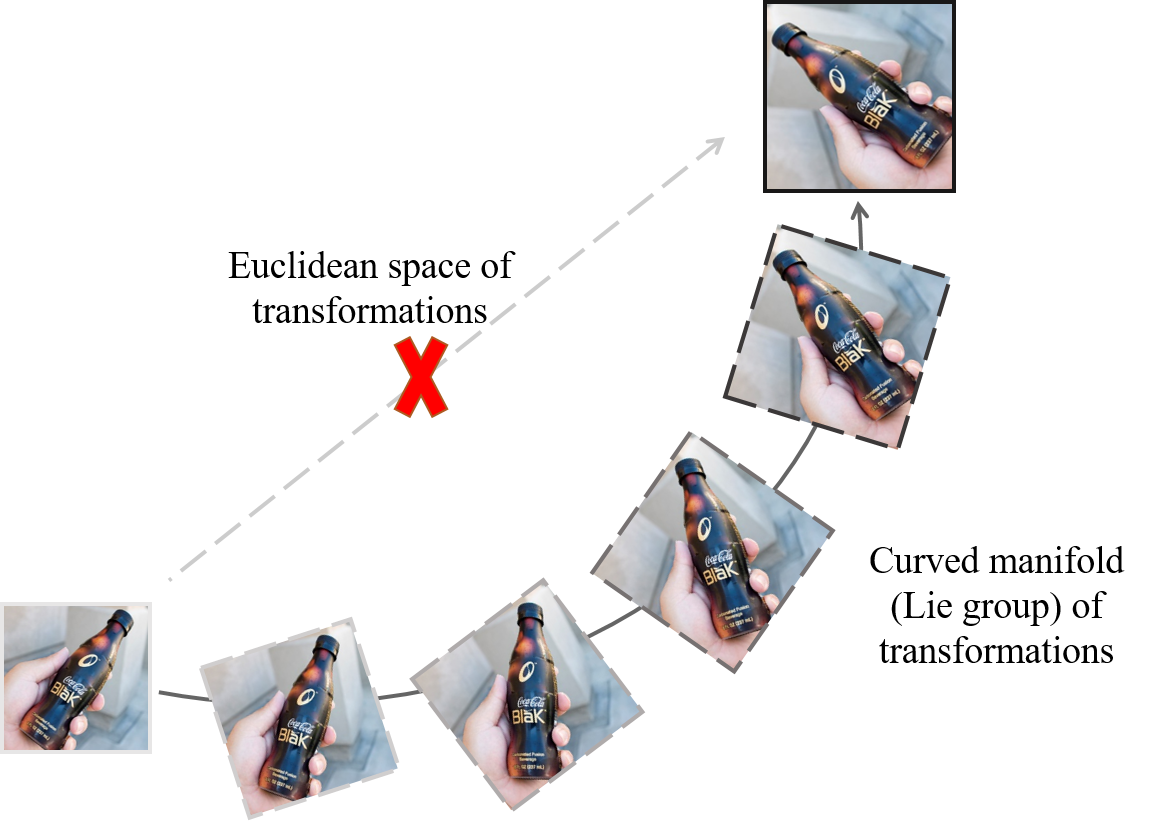}
        %\caption{AutoEncoding Data (AED)}
    \end{subfigure}\\\vspace{2mm}
    ~ %add desired spacing between images, e. g. ~, \quad, \qquad, \hfill etc.
      %(or a blank line to force the subfigure onto a new line)
    %\begin{subfigure}[c]{0.4\textwidth}
%        \includegraphics[width=\textwidth]{Figures/AET.png}
%        \caption{AutoEncoding Transformation (AET)}
%    \end{subfigure}\\\vspace{2mm}
%    \begin{subfigure}[c]{0.4\textwidth}
%        \includegraphics[width=\textwidth]{Figures/SAT.png}
%        \caption{(Semi-)Supervised Autoencoding Transformation (SAT)}
%    \end{subfigure}
    \caption{The deviation between two transformations should be measured along the curved manifold (Lie group) of transformations rather than through the forbidden Euclidean space of transformations. }\label{fig:manifold}
    %\vspace{-5mm}
\end{figure}

Learning powerful representations from data is one of the most significant topics in many deep learning problems. In particular, self-supervised representation learning from transformed images has shown great potential in both unsupervised and (semi-)supervised tasks, where the backbone networks are trained and regularized by decoding the self-supervisory signals without involving supervised labels \cite{qi2019avt,kolesnikov2019revisiting}.  These representations learned from transformations have also been well connected with the transformation equivariant representations as they change equivariantly in the same way as images are transformed.

The state-of-the-art performances have been demonstrated in literature for the transformation-based representation learning. Among them is the paradigm of AutoEncoding Transformations (AET) \cite{zhang2019aet,qi2019learning} that extends the family of autoencoders in a novel way. The idea is straightforward based on a simple criterion: a powerful representation should be able to capture the intrinsic visual structures of images before and after a transformation, such that the transformation can be well decoded from the representations of original and transformed images. In other words, the transformations are used as a critical weapon to reveal the intrinsic pattern of visual structures that change equivariantly to the applied transformations. Both deterministic and probabilistic AET models have been presented in literature, which minimizes the Mean-Square Errors (MSE) between the matrix representation of parametric transformations and maximizes the mutual information between the learned representation and transformations, respectively.

Although both AET models have shown the promising results, both the MSE and information maximization criteria suffer a questionable assumption that the Euclidean distance between transformation matrices is a good metric to characterize the difference between the estimation and the groundtruth. Unfortunately, this assumption is unrealistic. A group of transformations have a restricted form of the matrix representation on a manifold rather than scattering in the ambient Euclidean space of matrices. This suggests that a valid path connecting two transformations cannot be a straight line in the Euclidean space of transformation matrices. Instead it should be the shortest curve along the manifold of transformations, i.e., a geodesic.

This is illustrated in Figure~\ref{fig:manifold}. In other words, given a pair of original and transformed images, one cannot find a valid sequence of transformation matrices in the Euclidean space to continuously transform the original one to its transformed counterpart. Instead, a valid shortest path of transformations can only be sought on the manifold of the transformation group. This suggests that a more meaningful metric is the geodesic distance between two transformations defined along the underlying manifold.

Therefore, we resort to the theory of Lie group to study the geometry of transformations on the underlying manifold. We will formally present a new criterion to train the AET model by minimizing the geodesic distance between the estimated and applied transformations. In particular, we will use the homography group ${\bf PG}(2)$ to train the AET model, which contains a rich composition of image transformations such as rotations, translations and projective transformations.
While Riemannian logarithm is needed to characterize the geodesic distance between transformations, its calculation is intractable in the homography group without a closed form. To this end, we choose to project the transformations onto a subgroup with a tractable expression of geodesic distances.
In particular, we will show that the rotation group ${\bf SO}(3)$ is a subgroup of ${\bf PG}(2)$, where the geodesic distance and its derivative can be efficiently computed to train the model without having to explicitly take the logarithm. This can greatly reduce the computing overhead in self-training the proposed AETv2 model with the Lie group of homography transformations.

The remainder of this paper is organized as follows. In the next section, we will introduce some background about the transformation-based representation learning and point out its drawback.  Then, the criterion of geodesic distance minimization will be proposed to train the AETv2 in Section~\ref{sec:app}, following a brief discussion of  the preliminaries about the theory on the Lie group of transformations. We will discuss the implementation details about applying the homography group to train the proposed AETv2 model in Section~\ref{sec:imp}. Experiment results will be demonstrated in Section~\ref{sec:exp}, and we will conclude in Section~\ref{sec:concl}.
%------------------------------------------------------------------------
%\section{The Approach}

\section{Background and Notations}\label{sec:background}
First, let us review the previous work on Auto-Encoding Transformations (AET) \cite{zhang2019aet}, as well as discuss its drawback that will be addressed in this paper.

In the AET, a transformation $\mathbf t$ is sampled from a group $\mathcal G$, which is then applied to an image $\mathbf x$, resulting in a transformed copy $\mathbf t(\mathbf x)$. Usually, a transformation $\mathbf t$ can be represented by the corresponding matrix $\mathbf T$ in a group. For example, for the group ${\rm SIM}(2)$ of 2D similarity transformations, a transformation can be represented by a $3\times 3$ matrix $\mathbf T$ that is applied to a homogeneous space of coordinates to transform images spatially.

The goal of the AET is to learn a representation encoder $E(\mathbf x)$ for an image $\mathbf x$, as well as a transformation decoder $D$ to estimate the transformation $\mathbf t$ from the representations $E(\mathbf x)$ and $E(\mathbf t(\mathbf x))$ of original and transformed images.

It is supposed that a good representation encoder $E$ can capture the intrinsic visual structures of individual images, so that the decoder can infer the applied transformation by comparing the encoded representations of images before and after the transformations. Recent work has revealed the relation between the AET model and the transformation-equivariant representations from an information-theoretic point of view \cite{qi2019avt}.

Formally, to learn the encoder $E$ and the decoder $D$, the following Mean-Squared Error (MSE) is minimized to train the AET model with weights $\Theta$
\begin{equation}\label{eq:euclidean}
\bf \min_{\Theta} \mathop\mathbb E\limits_{\mathbf x,\mathbf t} {\rm\dfrac{1}{2}}\|{\hat  T}(\Theta) - T\|^2_{\rm F}
\end{equation}
where $\|\cdot\|_{\rm F}$ is the Frobenius norm of matrix, $\bf \hat T$ is an estimate of the matrix representation $\mathbf T$ of the sampled transformation $\mathbf t$, which is a function of the model parameters $\Theta$, and the mean-squared error is taken over the sampled images $\mathbf x$ and transformations $\mathbf t$.

Although the previous empirical results showed impressive performances of the learned AET representations, the above objective may not exactly characterize the distance between the estimate and the sampled transformation, as it simply uses the Euclidean distance between the matrix representations.

Indeed, a transformation group $\mathcal G$ is embedded into the ambient matrix space, thereby forming a manifold called a {\em Lie Group}. Obviously, a more accurate distance between two transformations should be characterized by the length of the geodesic connecting them, which characterizes how a transformation can  continuously change to another transformation along the underlying manifold. Minimizing such a geodesic distance on the manifold can yield a more exact estimate of a sampled transformation along the Lie group of transformations.

\section{The Proposed Approach}\label{sec:app}
In this section, we will briefly review the background knowledge about the Lie group in Section~\ref{sec:pre}. Then we will define the Geodesic Distance between Transformations (GDT) in Section~\ref{sec:gdt}, and present the training of the AET model by minimizing the GDT in Section~\ref{sec:aet_gdt}.

\subsection{Preliminaries}\label{sec:pre}

\begin{figure}[t]
%\captionsetup[subfigure]{singlelinecheck=false}
    \centering
    \begin{subfigure}[c]{0.5\textwidth}
        \includegraphics[width=\textwidth]{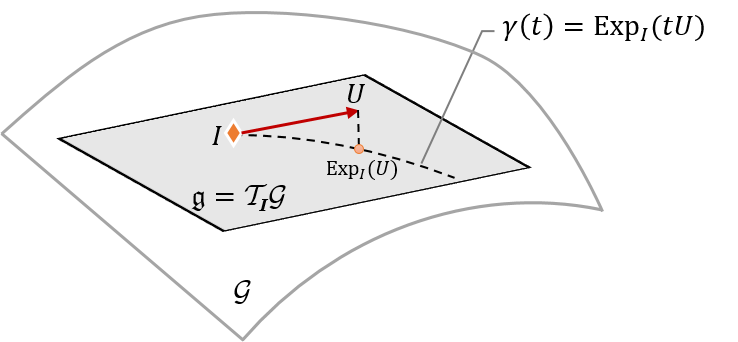}
        %\caption{AutoEncoding Data (AED)}
    \end{subfigure}\\\vspace{2mm}
    ~ %add desired spacing between images, e. g. ~, \quad, \qquad, \hfill etc.
      %(or a blank line to force the subfigure onto a new line)
    %\begin{subfigure}[c]{0.4\textwidth}
%        \includegraphics[width=\textwidth]{Figures/AET.png}
%        \caption{AutoEncoding Transformation (AET)}
%    \end{subfigure}\\\vspace{2mm}
%    \begin{subfigure}[c]{0.4\textwidth}
%        \includegraphics[width=\textwidth]{Figures/SAT.png}
%        \caption{(Semi-)Supervised Autoencoding Transformation (SAT)}
%    \end{subfigure}
    \caption{Illustration of a Lie group $\mathcal G$ of transformations with its Lie algebra $\mathfrak g$ as the tangent space $\bf \mathcal T_I \mathcal G$ at the identity $\mathbf I$. A geodesic curve $\rm \gamma(t)$ is defined by the Riemannian exponential $\bf {\rm Exp}_I (\mathrm t U)$ with the velocity $\mathbf U\in\mathfrak g$ at $\mathbf I$, where the Riemannian exponential $\bf {\rm Exp}_I$ maps $\bf U$ to the Lie group $\mathcal G$. }\label{fig:lie}
    %\vspace{-5mm}
\end{figure}

As illustrated in Figure~\ref{fig:lie}, a Lie group $\mathcal G$ of transformations is both a group and a manifold \cite{zacur2014left}. As a group, it is endowed with a composition $\circ$ and an inverse operator, which become the matrix product and inverse in the corresponding matrix representation.
For simplicity, we will use the matrix representation of transformations in the following exposition.

As a manifold, the tangent space $\mathcal T_\mathbf I\mathcal G$ at the identity transformation $\mathbf I$ plays a critical role, which is named the Lie algebra $\mathfrak g$ in literature \cite{zacur2014left}. It is a linear vector space endowed with the matrix addition and scalar multiplication. A group exponential $\exp$ is defined by mapping from the Lie algebra $\mathfrak g$ to the Lie group $\mathcal G$. In the matrix representation, the group exponential becomes matrix exponential \cite{zacur2014left}.

Besides the group exponential, when a Riemannian metric is defined on the manifold of Lie group, a geodesic $\gamma({\rm t}): [0,1] \rightarrow \mathcal G$ can be defined starting from the identity transformation $\gamma(0)=\mathbf I$ with an initial velocity $\gamma'(0)=\mathbf U\in \mathfrak g$. Then the Riemannian exponential is defined as $ {\rm Exp}_{\mathbf I}(\mathbf U)\triangleq \gamma(1)$. Note that the Riemannian exponential ${\rm Exp}_{\mathbf I}$ does not necessarily coincide with the group exponential unless there exists a bi-invariant metric \cite{zacur2014left} on the Lie group.

The group exponential may not be surjective, i.e., it does not map the Lie algebra $\mathfrak g$ {\em onto} the whole Lie group $\mathcal G$. In contrast, the Riemannian exponential is a surjective mapping onto a connected $\mathcal G$. This implies there exists a Riemannian logarithm ${\rm Log}_\mathbf I$ mapping from $\mathcal G$ back to $\mathfrak g$ such that $\bf {\rm Exp}_I({\rm Log}_I (U)) = U$.

Usually, we consider a left-invariant Riemannian metric $\bf \langle U, V\rangle_T$ on the Lie group $\mathcal G$ for $\bf U,V \in \mathcal T_T \mathcal G$, satisfying $\bf \langle U, V\rangle_T = \langle T^{-1}U, T^{-1}V\rangle_I$. Thus, one only needs to define an inner product on the Lie algebra $\mathfrak g$, and the inner product in the tangent space at the other transformations can be translated from that at the identity \cite{zacur2014left}. The left-invariant metric preserves the length of geodesics under a left translation by $\mathbf T$ (i.e., $\bf X\mapsto TX$ for each $\bf X\in\mathcal G$). Therefore, the Riemannian exponential ${\rm Exp}_\mathbf T$ at any transformation $\mathbf T$ can be written as $\bf {\rm Exp}_T(U)=T~{\rm Exp}_I(T^{-1}U)$ for $\bf U\in \mathcal T_T \mathcal G$.

\subsection{Geodesic Distance between Transformations}\label{sec:gdt}

In the AET, we seek to train the model by minimizing the error between the estimated and sampled transformations $\bf \hat T$ and $\bf T$. In the previous work, the mean-squared distance (aka the squared Euclidean distance) is adopted. However, it does not reflect the intrinsic distance between transformations along the manifold of the underlying Lie group. In contrast, it is only the external distance between the transformations in the ambient Euclidean space for matrices.

Thus, we instead use the geodesic to connect two transformations, which has the shortest length along the curved manifold between them. With the above discussion, it can be shown that the geodesic connecting $\bf \hat T$ and $\bf T$ can be represented as
$$
\bf \gamma ({\rm t})= {\rm Exp}_T ( {\rm t}~{\rm Log}_T (\hat T)) = T {\rm Exp}_I ( {\rm t}~{\rm Log}_I (T^{-1}\hat T))
$$
where $\gamma(0)=\mathbf T$ and $\gamma(1)=\bf \hat T$. Then, the geodesic distance between $\mathbf T$ and $\bf \hat T$ is defined as
$$
\rm \int_0^1 \|\gamma'( t)\| d t = \int_0^1 \sqrt{\langle \gamma'(t),\gamma'({\rm t})\rangle_{\gamma(t)}} d t= \|\gamma'(0)\|
$$
since $\gamma'(\rm t)$ is parallel along $\gamma(\rm t)$ and thus $\|\gamma'(\rm t)\|$ is a constant along the geodesic.

Therefore, we can minimize the following (squared) Geodesic Distance between Transformations (GDT) to train the AET model
\begin{equation}\label{eq:GDT}
\bf\ell(\hat T, T) \triangleq {\rm \dfrac{1}{2}} \|\gamma'({\rm 0})\|^\mathrm 2 = {\rm \dfrac{1}{2}}\|{\rm Log}_I (T^{-1}\hat T)\|^\mathrm 2
\end{equation}
where the last equality follows from the left invariance of the Riemannian metric.

\subsection{Autoencoding Transformations in Lie Groups}\label{sec:aet_gdt}
In the AET, $\bf \hat T$ in the loss (\ref{eq:GDT}) is the estimate of a sampled transformation $\bf T$ by the transformation decoder.
Let a residual matrix $\bf R$ be $\bf {\rm Log}_I (T^{-1}\hat T)$. Then, we can rewrite the geodesic distance between $\bf \hat T$ and $\mathbf T$ as
\begin{equation}\label{eq:loss}
\bf \ell(\hat T, T) = {\rm\dfrac{1}{2}} {\rm tr} (R^\intercal R)
\end{equation}
where
\begin{equation}\label{eq:exp_r}
\bf {\rm Exp}_I (R) = T^{-1}\hat T.
\end{equation}
Here $\mathbf R$ is viewed as a function of the estimation $\bf\hat T$ that further depends on the weights of the AET network.

In order to train the model via error back-propagation, we need to compute the derivative of $\bf \ell(\hat T, T)$ wrt $\bf\hat T$. By applying the chain rule of Fr$\acute{e}$chet matrix derivative wrt $\bf\hat T$ to LHS of (\ref{eq:exp_r}), we have
$$
\bf \nabla_{\hat T} {\rm Exp}_I (R) = \nabla_{R} {\rm Exp}_I (R) \nabla_{\hat T} R
$$

On the other hand, the derivative of the RHS of (\ref{eq:exp_r}) is
$$
\bf\nabla_{\hat T} (T^{-1}\hat T) = I \otimes T^{-1}.
$$

By equating the above two results, we obtain
$$
\bf \nabla_{\hat T} R =  (\nabla_{R} {\rm Exp}_I (R))^{-1} I \otimes T^{-1}
$$
where $\otimes$ is the Kronecker product, and $\bf I$ is the identity matrix with a compatible dimension in the matrix product.

By following the chain rule, one can write the derivative of the objective $\ell$ wrt the decoder output $\bf \hat T$
\begin{equation}\label{eq:grad}
\begin{aligned}
\bf \nabla_{\hat T} &\bf \ell(\hat T, T)   = \overline{R}^\intercal~\nabla_{\hat T} R \\
&\bf = \overline{R}^\intercal~(\nabla_{R} {\rm Exp}_I (R))^{-1} I \otimes T^{-1}
\end{aligned}
\end{equation}
where the overline $\bf \overline R$ denotes the stacking of the matrix columns. With the above derivative, the error signals will be back-propagated to update the network parameters.

In the next section, we will implement the AET with the group of homography transformations. We will discuss the challenge of directly computing its Riemannian exponential $\bf \bf{\rm Exp}_I$ and the corresponding (Fr$\acute{e}$chet) matrix derivative, and present an alternative tractable algorithm.
%------------------------------------------------------------------------
\section{The Implementation}\label{sec:imp}
In this section, we will discuss the implementation details about the proposed approach. First, we will review the Lie group of homography transformations in Section~\ref{sec:hom}, and show the challenge facing the direct application of Riemannian logarithm to compute the geodesic distance between transformations. Then, we will present an alternative algorithm in Section~\ref{sec:proj} by projecting the homography group onto a subgroup where there exists a tractable form of geodesic distance without explicitly computing the intractable Riemannian logarithm.

\subsection{Homography Transformations}\label{sec:hom}
The previous results have demonstrated that the group ${\rm\bf PG}(2)$ of homography transformations (aka projective transformations) have gained impressive performances with the AET model (referred to AETv1 in this paper) \cite{zhang2019aet}. It contains a rich family of spatial transformations that can reveal the visual structures of images.  Thus, we discuss the details about implementing the proposed AETv2 with ${\rm\bf PG}(2)$ below.

The 2D homography transformations ${\rm\bf PG}(2)$ can be defined as the transformations in the augmented 3D homogeneous space of image coordinates \cite{beutelspacher1998projective}. Its matrix representation contains all matrices with a unit determinant, and the corresponding Lie algebra $\mathfrak g$ consists of all matrices with zero trace.

With a left-invariant metric, the Riemannian exponential can be written in terms of matrix exponential \cite{zacur2014left},
$$
\bf {\rm Exp}_I (R) = \exp(R^\intercal) \exp(R-R^\intercal)
$$
for all $\bf R\in\mathfrak g$.  By applying the chain rule, the matrix derivative of $\bf {\rm Exp}_I$ can be rewritten as
\begin{equation}\label{eq:exp_grad}
\begin{aligned}
\bf \nabla_R & \bf {\rm Exp}_I (R) = \bf \exp(R^\intercal-R)\otimes I~{\rm d}\exp(R^\intercal) K \\
&\bf + I \otimes \exp(R^\intercal)~ {\rm d}\exp(R-R^\intercal) (I-K)
\end{aligned}
\end{equation}
where ${\rm d}\exp$ is the derivative of matrix exponential, and $\bf K$ is the communication matrix such that $\bf K \overline M = \overline {M^\intercal}$ for a matrix $\bf M$.

Then, substituting (\ref{eq:exp_grad}) in (\ref{eq:grad}) results in the derivative of the loss (\ref{eq:loss}) wrt the estimated transformation $\bf \hat T$, which can be back-propagated to train the model.

\subsection{Projection of ${\bf PG}(2)$ onto ${\bf SO}(3)$}\label{sec:proj}
The derivative in Eq.~(\ref{eq:grad}) involves the calculation of Riemannian logarithm $\mathbf R$ of $\bf T^{-1}\hat T$. Unfortunately, there is no closed-form solution to this Riemannian logarithm, and an optimization problem need be formulated to solve it by inverting the Riemannian exponential.
%\begin{equation}\label{eq:inv}
%\min_{R\in\mathfrak g} \ell_R (R) \triangleq \dfrac{1}{2}\|{\rm Exp}_I(R) - T^{-1}\hat T\|^2
%\end{equation}
%is often formulated to solve it.

%
%The matrices in this optimization problem usually has a much low dimension (e.g., $3\times 3$ matrices for ${\bf PG}(2)$), and thus the off-the-shelf optimizer like gradient decent methods can solve it very efficiently without negligible computing overhead.

Here we present an alternative method by projecting the target transformation $\bf T^{-1}\hat T$ onto a subgroup, where there exists a tractable Frobenius norm of Riemannian logarithm \cite{zacur2014multivariate}.
For the homography group ${\bf PG}(2)$, the rotation group ${\bf SO}(3)$ constitutes a subgroup by noting that its matrix representation consists of such $3\times 3$ matrices that $\bf T^\intercal T=I$ and $\det (\mathbf T) = 1$, and the corresponding Lie algebra contains all skew-symmetric matrices (i.e., $\bf R^\intercal = - R$) with traces equal to zero \cite{zacur2014left}. The Riemannian logarithm for the rotation subgroup is the matrix logarithm since the matrices commute in its Lie algebra.

Formally, an estimate of $\bf R$ can be made by projecting onto ${\bf SO}(3)$ and using the matrix logarithm in place of the Riemannian logarithm,
$$
\bf \hat R=\log\left[{\rm\mathop\Pi\limits_{{\bf SO}(3)}} (T^{-1}\hat T)\right]
$$
where $\log$ is the conventional matrix logarithm, and ${\rm\mathop\Pi\limits_{{\bf SO}(3)}} (\cdot)$ is the projection operator.

With the singular value decomposition of $\bf T^{-1}\hat T=U\Sigma V^\intercal$, it is not hard to show by the Karush-Kuhn-Tucker (KKT) conditions \cite{boyd2004convex} that the projection onto ${\bf SO}(3)$ can be written in a closed form as
$$
\bf P \triangleq {\rm\mathop\Pi\limits_{{\bf SO}(3)}} (T^{-1}\hat T) = U {\rm diag}\left\{1,1,\det(U V^\intercal)\right\} V^\intercal.
%\arg\min_{G\in {\bf SO(3)}}\|G-T^{-1}\hat T\|^2
$$

Moreover, by Rodrigues' rotation formula \cite{engo2001bch}, the Frobenius norm of the resultant $\bf \hat R \triangleq \bf \log P$ can be written as
$$
\bf \dfrac{\rm 1}{\rm 2}\|\hat R\|_{\rm F}^{\rm 2} \triangleq {\rm\dfrac{1}{2}}{\rm tr}(\hat R^\intercal \hat R) = \theta^{\rm 2},
$$
where
\begin{equation}\label{eq:theta}
\bf \theta=\arccos\left[\dfrac{{\rm tr} (P) - {\rm 1}}{\rm 2}\right]\in[{\rm 0},\pi]
\end{equation}
is the rotation angle around a unit 3D axis given by $\theta^{-1} \log \mathbf P$ whenever $\theta\neq 0$.

\begin{algorithm}[tb]
   \caption{Self-training the AETv2 with homography transformations}
   \label{alg:AETv2}
\begin{algorithmic}
   \STATE {\bfseries Input:} the model weights $\Theta$, minibatch size $n$
   \REPEAT
   \STATE Sample $n$ unlabeled images $\mathcal X=\{\mathbf x_i|i=1,\cdots,n\}$;
   \STATE Sample $n$ transformations $\mathcal T=\{\mathbf t_i|i=1,\cdots,n\}$, and apply to $\mathcal X$, resulting in the set of transformed images $\mathcal T(\mathcal X)=\{\mathbf t_i(\mathbf x_i)|i=1,\cdots,n\}$;
   \STATE Use the gradient of the loss (\ref{eq:appr_loss}) over the minibatch
   $$
    \dfrac{1}{n}\sum_{i=1}^n\nabla_\Theta\mathbf{\hat \ell} (\mathbf{\hat T}_i, \mathbf T_i)
   $$
   to update the model weights $\Theta$, with $\mathbf{\hat T}_i$ and $\mathbf T_i$ the matrix representation of the decoder output and the applied transformation for the $i$th sample.
   \UNTIL{Convergence}
\end{algorithmic}
\end{algorithm}

Then, we can approximate the loss $\bf \ell(\hat T, T)$ in Eq.~(\ref{eq:loss}) by combining the resultant geodesic distance and the distance between $\bf T^{-1}\hat T$ to its projection onto ${\bf SO}(3)$,
\begin{equation}\label{eq:appr_loss}
\bf \hat \ell(\hat T, T) = %{\rm tr}(\hat R^\intercal \hat R)
\arccos\left[\dfrac{{\rm tr} (P) - {\rm 1}}{\rm 2}\right] + \lambda~{\rm tr}(R_\Pi^\intercal R_\Pi)
\end{equation}
with the following projection residual
$$
\bf R_\Pi = T^{-1}\hat T - {\rm \mathop\Pi\limits_{{\bf SO}(3)}}(T^{-1}\hat T),
$$
where $\lambda$ is a positive weight on the projection distance, and it will be fixed to one in experiments.
Minimizing the projection residual can minimize the deviation incurred by projecting $\bf T^{-1}\hat T$ onto ${\bf SO}(3)$.

As illustrated in Figure~\ref{fig:AETv2}, we choose to minimize the above loss as a tractable surrogate to train the AET model without having to explicitly taking the matrix logarithm. Since most of common libraries automate the gradient of both the SVD and the determinant, the derivative of the projection can be implemented directly. Algorithm~\ref{alg:AETv2} summarizes the algorithm of self-training the AETv2 with homography transformations.

\section{Experiments}\label{sec:exp}

In this section, we present our experiment results by comparing the AETv2 with the  AETv1 as well as the other unsupervised models. Following the standard evaluation protocol in literature \cite{zhang2019aet,qi2019avt,oyallon2015deep,dosovitskiy2014discriminative,radford2015unsupervised,oyallon2017scaling,gidaris2018unsupervised}, we will adopt downstream classification tasks to evaluate the learned representations on CIFAR10, ImageNet, and Places datasets.

\subsection{Implementation Details}

\begin{figure}[t!]
%\captionsetup[subfigure]{singlelinecheck=false}
    \centering
    \begin{subfigure}[c]{0.46\textwidth}
        \includegraphics[width=\textwidth]{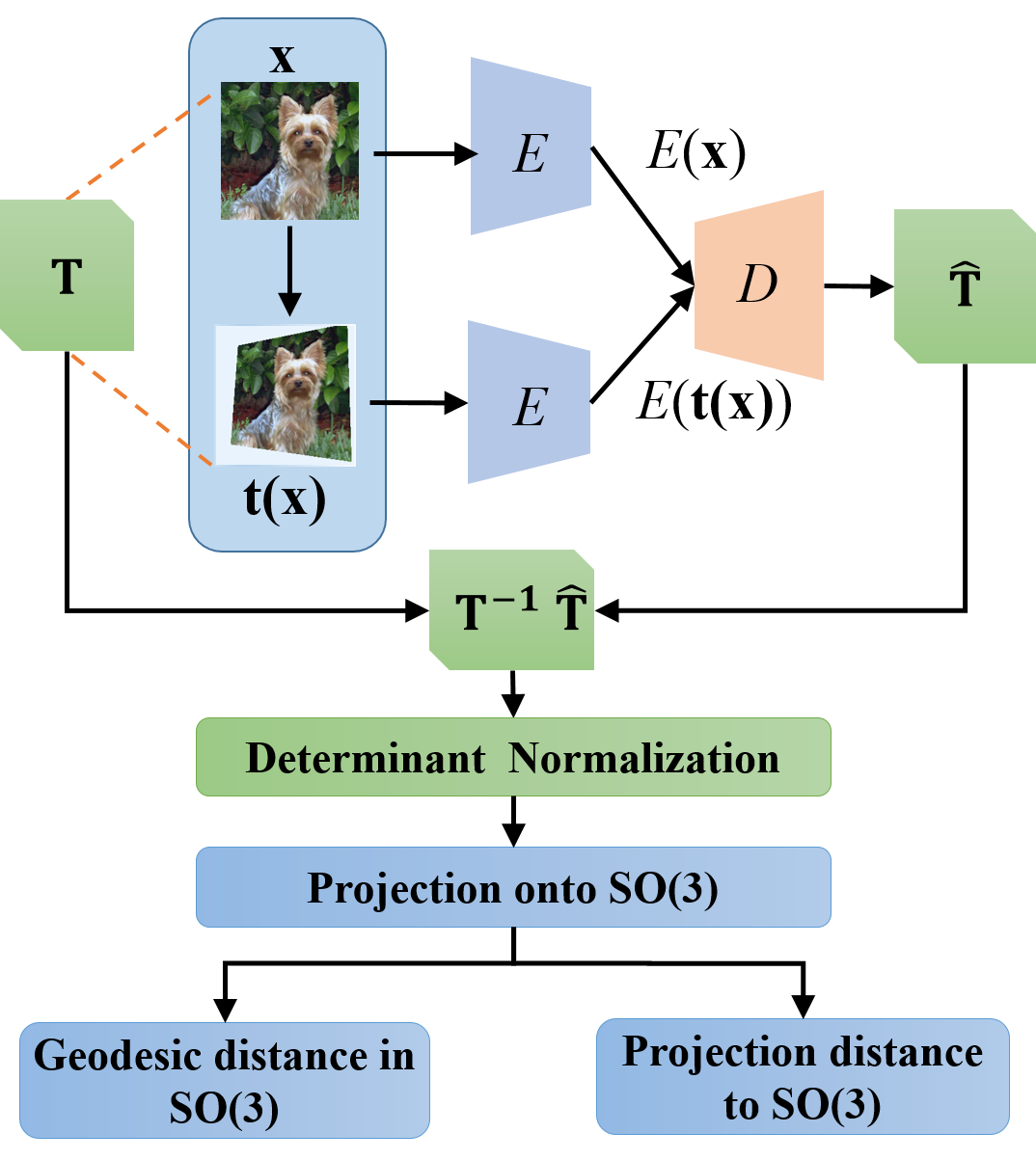}
        %\caption{AutoEncoding Data (AED)}
    \end{subfigure}\\\vspace{2mm}
    \caption{Illustration of how the AETv2 is self-trained end-to-end. The output matrix $\bf T^{-1}\hat{T}$ from the transformation decoder is normalized to have a unit determinant, and the projection of $\bf T^{\rm -1}\hat T$ onto ${\bf SO}(3)$ follows to compute the geodesic distance and the projection loss as in Eq.~(\ref{eq:appr_loss}) to train the model.}\label{fig:AETv2}
    %\vspace{-5mm}
\end{figure}

As in the AETv1 \cite{zhang2019aet}, a backbone network architecture is adopted to train a representation encoder and a transformation decoder for the AETv2 network. As shown in Figure~\ref{fig:AETv2}, an image and its transformed copy are fed into a Siamese encoder network, whose output representations are taken as the input into a transformation decoder. But unlike the AETv1, the AETv2 model works to decode the Lie group of homography transformations, and it outputs a matrix representation of the estimated transformations. Since a $3\times 3$ homography transformation matrix has eight degrees of freedom up to a scalar multiplier, its transformation matrix is obtained from the eight outputs by the decoder, with the last element fixed to one. The resultant transformation matrix $\bf T^{-1} \hat T$ is finally normalized to have a unit determinant as illustrated in Figure~\ref{fig:AETv2}. %The pipeline is summarized that shows how the AETv2 model is self-trained end-to-end.

For the sake of a fair comparison, a homography transformation is sampled like in the AETv1 by
randomly translating four corners of an image in both horizontal and vertical directions by $\pm 0.125$ of its height and width,  after it is randomly scaled by $[0.8, 1.2]$ and rotated by $0^\circ, 90^\circ, 180^\circ,$ or $270^\circ$.

\subsection{CIFAR-10 Experiments}

\subsubsection{Network and Experiment Details}
To make a fair and direct comparison with existing unsupervised models, we adopt the Network-In-Network (NIN) architecture previously used on the CIFAR-10 dataset for the unsupervised learning task \cite{zhang2019aet,qi2019avt,oyallon2015deep,dosovitskiy2014discriminative,radford2015unsupervised,oyallon2017scaling,gidaris2018unsupervised}.  The NIN consists of four convolutional blocks, each having three convolutional layers.
Two Siamese NIN branches are then constructed in the AETv2, each of which takes the original and the transformed images, respectively. The outputs from the last block of two branches are concatenated and average-pooled to form a $384$-d feature vector. An output layer follows to predict the eight parameters of the input homography transformation.
The model is trained by the Adam solver with a learning rate of $10^{-5}$, a value of $0.9$ and $0.999$ for $\beta_1$ and $\beta_2$, and a weight decay rate of $5\times 10^{-4}$.
%and the weight on the projection loss are set to $5\times 10^{-4}$ and one, respectively.

A classifier is then built on top of the second convolutional block to evaluate the quality of the learned unsupervised representation following the standard protocol in literature \cite{zhang2019aet,qi2019avt,oyallon2015deep,dosovitskiy2014discriminative,radford2015unsupervised,oyallon2017scaling,gidaris2018unsupervised}.
In particular, the first two blocks are frozen when the classifier atop is trained with labeled examples.

Both model-based and model-free classifiers are trained for the evaluation purpose. First, we train a non-linear classifier with various numbers of Fully-Connected (FC) layers. Each hidden layer has $200$ neurons followed by a batch-normalization and ReLU activation. We also train a convolutional classifier by adding a third NIN block  on top of the unsupervised features, and its output feature map is averaged pooled and connected to a linear softmax layer.

Alternatively, we test a model-free KNN classifier based on the averaged-pooled features from the second convolutional block. Without explicitly training a model with labeled data, the KNN classifier can make a direct assessment on the quality of unsupervised feature representations.

%The two branches share the same network weights, and are used as the encoder network producing the feature representations for input images.

\subsubsection{Results}

\begin{table}
\caption{Comparison between unsupervised models on CIFAR-10. The fully supervised NIN and the random Init. + conv have the same three NIN blocks, but they are fully supervised and is trained with the first two blocks randomly initialized and staying frozen during training, respectively. ``+FC" and ``+conv" denote a nonlinear classifier with a hidden FC layer and a NIN convolutional block respectively, followed by a 10-way softmax layer. }\label{tab:cifar10:comp01}
\centering\small
 \begin{tabular}{l|c} \toprule
Method&Error rate\\ \midrule
Supervised NIN (Lower Bound)&7.20  \\
Random Init. + conv (Upper Bound)&27.50  \\ \midrule
Roto-Scat + SVM \cite{oyallon2015deep} &17.7 \\
ExamplarCNN \cite{dosovitskiy2014discriminative} &15.7 \\
DCGAN \cite{radford2015unsupervised}&17.2 \\
Scattering \cite{oyallon2017scaling}&15.3\\
RotNet + FC \cite{gidaris2018unsupervised}&10.94\\
RotNet + conv \cite{gidaris2018unsupervised}&8.84\\
%(Ours) AET-affine + FC &9.77\\
%(Ours) AET-affine + conv &8.05\\
AETv1 + FC \cite{zhang2019aet} &9.41\\
AETv1 + conv \cite{zhang2019aet} &7.82\\ \midrule
(Ours) AETv2 + FC & \bf 9.09\\
(Ours) AETv2 + conv & \bf 7.44\\\bottomrule
\end{tabular}
\end{table}

\begin{table*}
\caption{Comparison between AETv1 and AETv2 on CIFAR-10 with different classifiers on top of learned representations for evaluation, where $n$-FC denotes a $n$-layer fully connected (FC) classifier, and the KNN is obtained with $K=10$ nearest neighbors. The numbers in parentheses are the {\it relative} reduction in error rates w.r.t. the AETv1 baseline.}\label{tab:cifar10:comp02}
\centering\small
 \begin{tabular}{c|ccccc} \toprule
   &KNN&1-FC&2-FC&3-FC&conv\\ \midrule
%RotNet baseline~\cite{gidaris2018unsupervised}&24.97 &18.21&11.34&10.94 &8.84 \\
AETv1 \cite{zhang2019aet}&22.39 &16.65 &9.41 &9.92 &7.82\\ \
(Ours) AETv2 &\textbf{21.26} ($\downarrow$ 5.0\%)&\textbf{15.03} ($\downarrow$ 9.7\%)&\textbf{9.09} ($\downarrow$ 3.4\%)&\textbf{9.55} ($\downarrow$ 3.7\%)&\textbf{7.44} ($\downarrow$ 4.9\%) \\ \bottomrule
\end{tabular}
\end{table*}

Table~\ref{tab:cifar10:comp01} compares the AETv2 with the other models on CIFAR-10.
On one hand, it outperforms the AETv1 as well as the other unsupervised models with the same backbone. Furthermore, it narrows the performance gap with the fully supervised convolutional classifier ($7.44\%$ vs. $7.20\%$) that gives the lower bound of error rate when all labels are used to train the model end-to-end.

More comparisons with the AETv1 are made in Table~\ref{tab:cifar10:comp02}. We compare the performances by both the model-based and model-free classifiers in the downstream tasks for both versions of the AET models. From the results, we can see that the AETv2 consistently outperforms its AETv1 counterpart with both KNN classifiers and different numbers of fully connected layers.
\subsection{ImageNet Experiments}

\subsubsection{Network and Experiment Details}
We further evaluate the performance on the ImageNet dataset. For a fair comparison with the AETv1 \cite{zhang2019aet}, the AlexNet is used as the backbone to learn the unsupervised features.  Two branches with shared parameters are created by taking original and transformed images as inputs to train the unsupervised model. The $4,096$-d output features from the second last fully connected layer in two branches are concatenated and fed into the output layer producing eight projective transformation parameters. We still use the Adam solver to train the network with a batch size of $768$ original and transformed images. %while the weight on the projection loss is fixed to two on ImageNet. %The initial learning rate is set to $0.01$, and it is dropped by a factor of $10$ at epoch 100 and 150. AET is trained for $200$ epochs in total.

%the homography transformations applied are randomly sampled in the same fashion as on CIFAR-10.
\subsubsection{Results}

\begin{table}
\caption{Top-1 accuracy on ImageNet.  After unsupervised training of AlexNet, nonlinear classifiers are trained on top of Conv4 and Conv5 layers with labeled data for the evaluation purpose. The fully supervised and random models are also compared that give upper and lower bounded performances, respectively. A single crop is applied with no dropout or local response normalization during the testing. }\label{tab:imagenet:01}
%* The result by DeepCluster is based on ten random crops of images during the testing.}
\centering\small
 \begin{tabular}{l|cc} \toprule
Method&Conv4 &Conv5\\ \midrule
ImageNet Labels \cite{bojanowski2017unsupervised}(Upper Bound)&59.7&59.7  \\
Random \cite{noroozi2017representation} (Lower Bound)&27.1 &12.0  \\ \midrule
Tracking \cite{wang2015unsupervised} &38.8&29.8 \\
Context \cite{doersch2015unsupervised} &45.6&30.4 \\
Colorization \cite{zhang2016colorful}&40.7&35.2 \\
Jigsaw Puzzles \cite{noroozi2016unsupervised}&45.3&34.6\\
BiGAN \cite{donahue2016adversarial}&41.9&32.2\\
NAT \cite{bojanowski2017unsupervised}&-&36.0\\
DeepCluster \cite{caron2018deep} &-&44.0\\
RotNet \cite{gidaris2018unsupervised}&50.0&43.8\\
AETv1 \cite{zhang2019aet} &{53.2}&{47.0}\\\midrule
(Ours) AETv2 &\textbf{54.3}&\textbf{47.5}\\\bottomrule
\end{tabular}
\end{table}

\begin{table*}
\caption{Top-1 accuracy on ImageNet, where a $1,000$-way linear classifier is trained on top of different convolutional layers of feature maps spatially resized to about $9,000$ elements. Fully supervised and random models show the upper and the lower bounds of unsupervised model performances. Only a single crop is used during testing, except that the models marked with ``*" apply ten crops.}\label{tab:imagenet:02}
\centering\small
 \begin{tabular}{l|ccccc} \toprule
Method&Conv1 &Conv2&Conv3&Conv4&Conv5\\ \midrule
ImageNet Labels (Upper Bound) \cite{gidaris2018unsupervised}&19.3&36.3&44.2&48.3&50.5  \\
Random (Lower Bound)\cite{gidaris2018unsupervised} &11.6 &17.1&16.9&16.3&14.1  \\
Random rescaled \cite{krahenbuhl2015data}(Lower Bound)&17.5 &23.0&24.5&23.2&20.6  \\\midrule
Context \cite{doersch2015unsupervised} &16.2&23.3&30.2&31.7&29.6 \\
Context Encoders \cite{pathak2016context}&14.1&20.7&21.0&19.8&15.5 \\
Colorization\cite{zhang2016colorful}&12.5&24.5&30.4&31.5&30.3\\
Jigsaw Puzzles \cite{noroozi2016unsupervised}&18.2&28.8&34.0&33.9&27.1\\
BiGAN \cite{donahue2016adversarial}&17.7&24.5&31.0&29.9&28.0\\
Split-Brain \cite{zhang2017split}&17.7&29.3&35.4&35.2&32.8\\
Counting \cite{noroozi2017representation}&18.0&30.6&34.3&32.5&25.7\\
RotNet \cite{gidaris2018unsupervised}&18.8&31.7&38.7&38.2&36.5\\
AETv1 \cite{zhang2019aet} &{19.2}&{32.8}&{40.6}&{39.7}&{37.7}\\\midrule
(Ours) AETv2 &\textbf{19.6}&\textbf{34.1}&\textbf{41.9}&\textbf{40.4}&\textbf{37.9}\\\midrule\midrule
DeepCluster* \cite{caron2018deep} &13.4&32.3&41.0&39.6&38.2\\
AETv1* \cite{zhang2019aet} &{19.3}&{35.4}&{44.0}&{43.6}&{42.4}\\\midrule
(Ours) AETv2* &\textbf{21.2}&\textbf{36.9}&\textbf{45.9}&\textbf{44.7}&\textbf{43.2}\\
\bottomrule
\end{tabular}\\
\end{table*}

Table~\ref{tab:imagenet:01} reports the Top-1 accuracies of compared methods on ImageNet with the evaluation protocol used in \cite{noroozi2016unsupervised}, where Conv4 and Conv5 denote the training of AlexNet with the labeled data, after the bottom convolutional layers up to Conv4 and Conv5 are pre-trained in an unsupervised fashion and frozen thereafter.  The results show that in both settings, the AETv2 outperforms the other unsupervised models including the AETv1. The performance gap to the fully supervised models that give the upper bounded performance has been further narrowed to $5.4\%$ and $12.2\%$.

To evaluate the quality of unsupervised representations, a weak $1,000$-way fully connected linear classifier is trained on top of different numbers of convolutional layers. The results are shown in Table~\ref{tab:imagenet:02}, and the AETv2 again achieves the best Top-1 accuracy among the compared models. This shows that the AETv2 can learn a high-quality unsupervised representation with superior performances even though a weaker classifier is used.
%While the fully supervised models give the upper bounded performance by training the entire network with all labeled data, The result also demonstrates that the AETv2 further narrows the performance gap with the fully supervised models. The gap to the fully supervised Top-1 accuracy decreases from $6.5\%$ and $12.7$
%
% The classifiers of random models are trained on top of Conv4 and Conv5 with randomly sampled weights, and they set up the lower bounded performance. From the comparison, the AET models greatly narrow the performance gap to the upper bound -- the gap to the upper bound Top-1 accuracy has been decreased from $9.7\%$ and $15.7\%$ by RotNet and DeepCluster on Conv4 and Conv5, respectively, to $6.5\%$ and $12.7\%$ by AET, which is {\it relatively} narrowed by $33\%$ and $19\%$,  respectively.

%Moreover, we also follow the testing protocol adopted in \cite{zhang2017split} to compare the models by training a $1,000$-way linear classifier on top of different numbers of convolutional layers in Table~\ref{tab05}.  Again, AET obtains the best accuracy among all the compared unsupervised models.

\subsection{Places Experiments}

We also conduct experiments to evaluate unsupervised models on the Places dataset.
An unsupervised  representation is first pretrained on the ImageNet, and a single-layer softmax classifier is trained on top of different layers of the feature maps with Places labels. In this way, we assess how well unsupervised features can generalize across datasets. As shown in Table~\ref{tab:places},  the AETv2 outperforms the compared unsupervised models again, except for Conv1 and Conv2 in which case Counting \cite{noroozi2017representation} performs slightly better.

\begin{table*}
\caption{Top-1 accuracy on the Places dataset. Different layers of feature maps are spatially resized to about $9,000$ elements, and a $205$-way linear classifier is trained atop. All unsupervised features are pre-trained on the ImageNet, and are frozen when training the classification layer with Places labels.  The fully-supervised networks trained with Places Labels and ImageNet labels are also compared, in addition to random models. The best accuracies are highlighted in bold and the second best values are underlined. }\label{tab:places}
\centering\small
 \begin{tabular}{l|ccccc} \toprule
Method&Conv1 &Conv2&Conv3&Conv4&Conv5\\ \midrule
Places labels \cite{zhou2014learning}&22.1&35.1&40.2&43.3&44.6 \\
ImageNet labels&22.7&34.8&38.4&39.4&38.7\\
Random &15.7 &20.3&19.8&19.1&17.5  \\
Random rescaled \cite{krahenbuhl2015data}&21.4 &26.2&27.1&26.1&24.0  \\
\midrule
Context \cite{doersch2015unsupervised} &19.7&26.7&31.9&32.7&30.9 \\
Context Encoders \cite{pathak2016context}&18.2&23.2&23.4&21.9&18.4 \\
Colorization\cite{zhang2016colorful}&16.0&25.7&29.6&30.3&29.7\\
Jigsaw Puzzles \cite{noroozi2016unsupervised}&\underline{23.0}&31.9&35.0&34.2&29.3\\
BiGAN \cite{donahue2016adversarial}&22.0&28.7&31.8&31.3&29.7\\
Split-Brain \cite{zhang2017split}&21.3&30.7&34.0&34.1&32.5\\
Counting \cite{noroozi2017representation}&\textbf{23.3}&\textbf{33.9}&{36.3}&{34.7}&29.6\\
RotNet \cite{gidaris2018unsupervised}&21.5&31.0&35.1&34.6&{33.7}\\
%DeepCluster* \cite{} &19.6&33.2&39.2&39.8&34.7\\
AETv1 \cite{zhang2019aet} &22.1&{32.9}&\underline{37.1}&\underline{36.2}&\underline{34.7}\\\midrule
AETv2 &22.8&\underline{33.2}&\textbf{38.1}&\textbf{36.8}&\textbf{35.3}\\\bottomrule
\end{tabular}
\end{table*}

\section{Conclusions}\label{sec:concl}
In this paper, we present a novel AETv2 model by minimizing the geodesic distances between transformations along the manifold   instead of the mean-squared errors in the ambient Euclidean space of matrices as before.  While the geodesic characterizes how a transformation continuously evolves to another one in the Lie group of transformations, the AETv2 model can be self-trained to learn a better unsupervised representation by minimizing the deviation between the predicted and the applied transformations. In particular, we implement the AETv2 on the manifold of homography transformations, a Lie group containing rich spatial transformations to reveal the intrinsic visual structures.  However, an intractable Riemannian logarithm has to be computed to solve the geodesics between homography transformations. Alternatively, we choose to project the decoded transformations onto the ${\bf SO}(3)$ subgroup, and train the AETv2 by minimizing a tractable geodesic distance in ${\bf SO}(3)$ along with the associated projection loss. Experiment results demonstrate the AETv2 greatly outperforms the previous version as well as the other unsupervised models on several datasets.

{\small
\bibliographystyle{ieee_fullname}
\bibliography{ATLGT,AET}
}

\end{document}